\documentclass{article}
\usepackage{microtype}
\usepackage{graphicx}
\usepackage{subcaption}
\usepackage{booktabs}
\usepackage{amsmath,amssymb,mathtools,amsthm}
\usepackage{xurl} 
\usepackage{hyperref}

\usepackage[preprint]{icml2027}
\usepackage{currfile}
\usepackage{placeins}
\usepackage[capitalize,noabbrev]{cleveref}

\icmltitlerunning{AeroJEPA: Aerosol Forecasting with Predictive Learning}

\newcommand{\papertitle}{AeroJEPA: A Two-Stage Framework for Aerosol Forecasting with Joint-Embedding Predictive Learning}
\begin{document}
\twocolumn[
  \icmltitle{\papertitle}
  \begin{icmlauthorlist}
    \icmlauthor{Zitong Shuai}{seu}
    \icmlauthor{Zengliang Zang}{nudt}
    \icmlauthor{Chunlei Shi}{seu}
    \icmlauthor{Lang Liu}{nudt}
    \icmlauthor{Ning Liu}{nudt}
    \icmlauthor{Yiwen Hu}{nudt}
    \icmlauthor{Dan Niu}{seu}
    \icmlauthor{Chenyan Zhou}{seu}
    \icmlauthor{Shangjian Xie}{seu}
  \end{icmlauthorlist}
  \icmlaffiliation{seu}{School of Automation, Southeast University}
  \icmlaffiliation{nudt}{College of Meteorology and Oceanography, National University of Defense Technology}
  \icmlcorrespondingauthor{Dan Niu}{danniu1@163.com}
  \icmlkeywords{Aerosol forecasting, joint-embedding predictive learning, spatiotemporal prediction}
  \vskip 0.3in
]
\printAffiliationsAndNotice{}

\begin{abstract}
Aerosol forecasting requires integrating sparse ground observations and
modeling meteorologically driven concentration evolution and spatial
structure. However, analysis-initialized forecasts make limited explicit use
of historical observations, while pointwise regression tends to smooth local
variations and provides limited constraints on fine spatial structures. In
this work, we propose AeroJEPA, a two-stage joint-embedding predictive
framework for regional surface PM$_{2.5}$ forecasting. Specifically, Stage I
fuses historical observations, model backgrounds, and meteorological
information to predict an initial future state, meteorological conditions, and
multi-scale structures. Morphology-guided encoding and joint-embedding
predictive supervision support learning future spatial representations,
supplementing concentration regression with structural constraints. Stage II
uses predicted meteorology to guide state refinement and incorporates coarse-
and fine-scale structures into state updates and decoding to generate the
concentration field at a single future time step. Experiments show that
AeroJEPA reduces gridded and independent-station RMSE by 17.23\% and 7.90\%,
respectively, relative to PFGNet, while improving spatial consistency with
analysis fields. In a 48 h recursive forecasting case study, AeroJEPA
generally achieves lower RMSE and MAE than the numerical forecasts across the
forecast period, indicating promising short- to medium-range forecasting
performance.

\end{abstract}

\begin{figure}[t]
  \centering
  \includegraphics[width=\linewidth]{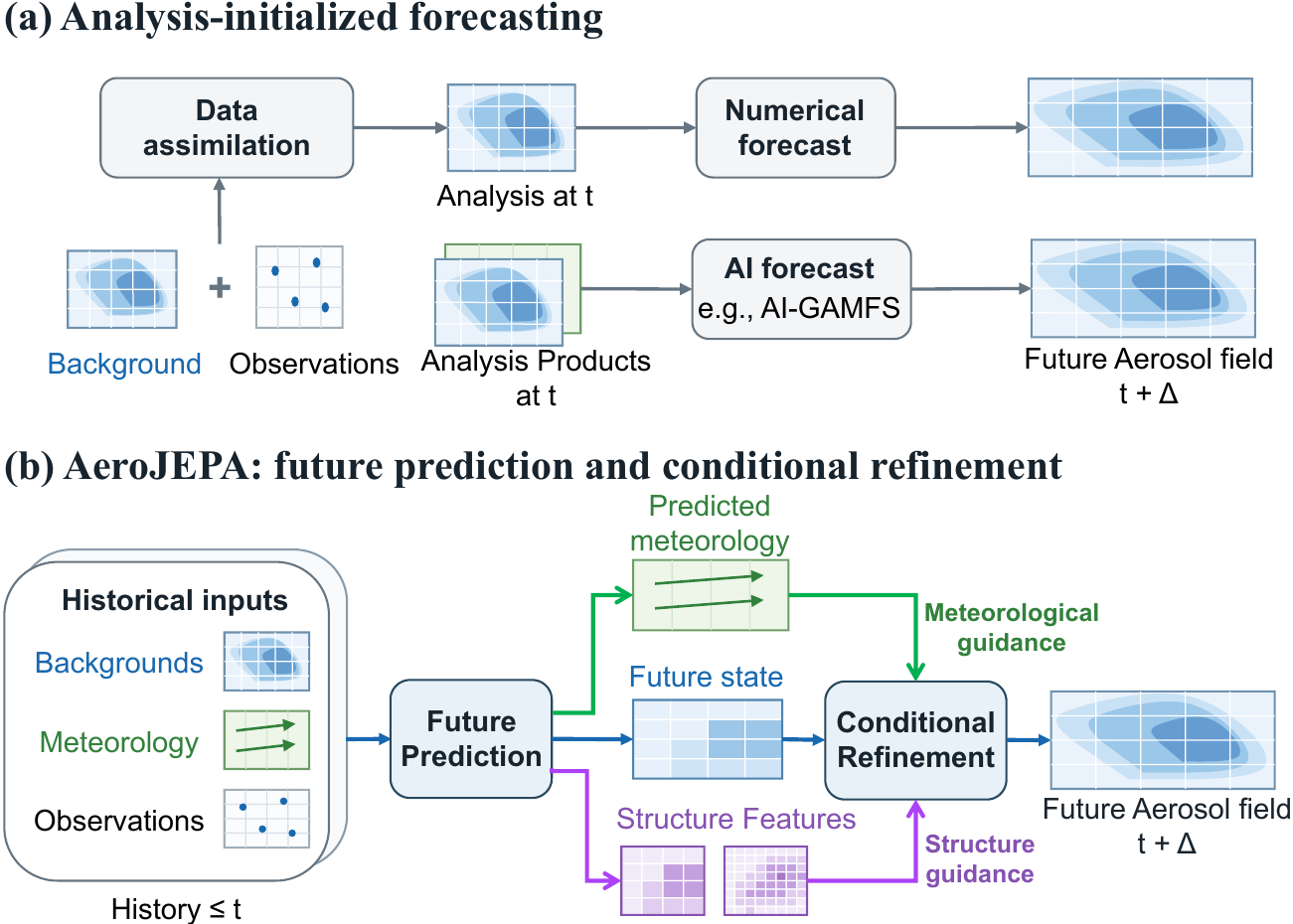}
  \caption{(a) Analysis-initialized numerical and AI forecasting.
  (b) AeroJEPA combines historical observations and model information for
  concentration field forecasting through future prediction and conditional
  refinement guided by meteorology and spatial structures.}
  \label{fig:observation_pathways}
\end{figure}

\section{Introduction}
\label{sec:introduction}
Reliable forecasts of particulate matter concentrations are essential for air quality management and early warning of pollution episodes~\citep{gui2026aigamfs}. At its core, aerosol concentration forecasting is a meteorologically driven spatiotemporal prediction problem involving multiple coupled variables: future concentration fields depend on both historical concentration states and the combined effects of pollutant transport, dispersion, turbulent mixing, and chemical transformation~\citep{liang2023airformer,han2021mastergnn,bodnar2025aurora}. Changes in winds, temperature, and boundary-layer structure can cause similar historical concentration states to follow different evolutionary trajectories. Jointly modeling the spatiotemporal evolution of meteorological conditions and aerosol concentrations is therefore essential for accurate forecasting.

Air-quality forecasting has developed along two main lines: physics-based
numerical modeling and data-driven learning. Numerical methods are based on
atmospheric physics and chemistry and may incorporate observations through
data assimilation~\citep{grell2005wrfchem,ha2022aerosolda,bocquet2015data}.
Data-driven methods learn spatiotemporal dependencies in pollutant
evolution~\citep{liang2023airformer,han2021mastergnn}, with some approaches
further incorporating physical guidance or observation
fusion~\citep{hettige2024airphynet,tian2025airdualode,cai2024convlstmenkf}.
These methods cover station-level concentration prediction and spatially
continuous gridded-field forecasting. The latter encompasses global
atmospheric-composition forecasting~\citep{bodnar2025aurora,gui2026aigamfs} and
regional long-horizon
forecasting~\citep{zhu2025interpretablepm25,pang2026zeeman,kang2026fakerair}.


However, accurate prediction of future concentration fields still presents three
challenges. First, as shown in \cref{fig:observation_pathways}(a), analysis-initialized
forecasts use observations indirectly through analysis products, without taking
historical observations and model backgrounds as separate inputs to the predictor.
Second, beyond jointly forecasting meteorology and aerosols, how meteorological
features influence aerosol concentrations remains underexplored. Finally, when
future evolution is uncertain, deterministic predictors trained primarily with
pointwise regression objectives such as MSE tend to produce overly smooth fields,
attenuating local variations and fine spatial structures~\citep{mathieu2016deep,yu2024diffcast}.
Thus, concentration regression should be complemented by more explicit learning
constraints on future spatial structures.


To address these challenges, we propose AeroJEPA, a two-stage joint-embedding
predictive framework for regional surface PM$_{2.5}$ forecasting. The framework
first predicts an initial future latent state together with meteorological
conditions and spatial structures at the target time, then uses these predicted
conditions to refine the concentration field. As shown in
\cref{fig:observation_pathways}(b), Stage I explicitly integrates historical
ground observations with model backgrounds and meteorology. Observations,
observation--background differences, and validity masks inform updates to
historical features, allowing historical observations to participate directly
in future-state prediction. To account for meteorologically driven concentration
evolution, Stage II uses the predicted meteorological fields and their latent
representations as refinement conditions. Feature interactions across
meteorological variables and local neighborhoods allow each location to select
relevant information, while latent meteorological features modulate subsequent
state updates, giving meteorological prediction a direct role in constructing
the future concentration field beyond auxiliary supervision. To complement
pointwise concentration regression with spatial structural constraints, we
further introduce morphology-guided encoding and joint-embedding predictive
supervision. The encoder combines full-field spatial features with local
morphology-guided enhancement to represent both broad concentration patterns
and local variations. During training, multi-scale representations of future
analysis fields serve as structure prediction targets, explicitly supervising
spatial organization at the forecast time. The predicted coarse structures
are incorporated into future-state updates, while fine structures are fused
with the refined state before decoding, allowing structure learning to guide
both latent-state evolution and final concentration field generation. Together,
these designs connect historical observation fusion, future meteorological
guidance, and multi-scale structure prediction to support future concentration
field forecasting.


Our main contributions are as follows:
\begin{itemize}
    \item We propose a two-stage framework for regional aerosol forecasting
    that integrates historical observations and meteorological guidance.
    The framework explicitly combines historical ground observations, model
    backgrounds, and meteorology, and uses predicted meteorological fields
    and their latent representations as conditions for refining the future
    concentration state, allowing meteorological predictions to directly
    inform concentration field generation.
    \item We design a mechanism for learning and fusing future multi-scale
    spatial structures. Morphology-guided encoding and joint-embedding
    predictive supervision enable learning of future coarse- and fine-scale
    structures, which are incorporated into state updates and decoding,
    respectively, to provide additional spatial structural constraints
    for concentration regression.
    \item We evaluate forecasting performance at both the gridded-field and
    independent-station levels, and examine the effectiveness of key designs
    through ablation experiments. We further conduct an extended recursive
    forecasting case study to assess the single-step model under repeated
    prediction feedback without new observations.
\end{itemize}




\section{Related work}
\label{sec:related_work}

\subsection{Spatiotemporal Forecasting}
\label{sec:related_spatiotemporal}

\textbf{Spatiotemporal forecasting models} capture spatial relationships and
temporal dependencies to predict future spatial fields from historical
sequences. Representative methods include convolution-based
SimVP~\citep{gao2022simvp,tan2025simvpv2}, attention-based
Earthformer~\citep{gao2022earthformer}, and AFNO~\citep{guibas2022afno},
which mixes information in the Fourier domain.
PFGNet~\citep{cai2026pfgnet} further introduces frequency-guided gating
to adaptively fuse multi-scale responses. For air quality forecasting,
AirFormer~\citep{liang2023airformer} learns dependencies across monitoring
stations, while AirPhyNet~\citep{hettige2024airphynet} and
Air-DualODE~\citep{tian2025airdualode} incorporate atmospheric physical
processes into concentration dynamics. Beyond station-level prediction,
Aurora~\citep{bodnar2025aurora} and AI-GAMFS~\citep{gui2026aigamfs}
forecast atmospheric composition on global grids, whereas
FAKER-Air~\citep{kang2026fakerair} focuses on regional forecasting over
long horizons and the reliability of operational warnings. These studies
span station-level concentration prediction and continuous gridded-field
forecasting, offering different modeling approaches through spatiotemporal
dependency learning and physical guidance.


\textbf{Observations and meteorology} offer additional avenues for improving
concentration forecasts. ConvLSTM-EnKF~\citep{cai2024convlstmenkf}
combines data assimilation with data-driven prediction, incorporating new
observations to correct subsequent forecasts. Joint forecasting exploits
relationships among atmospheric variables by learning their evolution
together: MasterGNN~\citep{han2021mastergnn} jointly models air quality
and weather, while AI-GAMFS~\citep{gui2026aigamfs} also predicts aerosols
and meteorological variables together. Beyond joint supervision,
DAG~\citep{qiu2026dag} studies time series forecasting with exogenous
variables through the learning and reuse of temporal and cross-variable
dependencies, demonstrating the value of using relationships learned from
auxiliary variables for target prediction. These approaches represent
different ways of using information, including observational correction,
joint training, and relationship reuse. We further investigate the explicit
fusion of historical observations with model backgrounds and the role of
meteorological features in guiding future concentration-state updates.


\subsection{Predictive Representation Learning}
\label{sec:related_predictive_representation}

Joint-embedding predictive learning predicts target representations from
context to supervise intermediate representations. I-JEPA~\citep{assran2023ijepa}
predicts features of target image regions from image context, constructing
prediction tasks in representation space. V-JEPA~\citep{bardes2024vjepa}
extends this approach to video, learning video representations by predicting
the spatiotemporal representations of masked regions. Furthermore,
V-JEPA 2-AC, the action-conditioned extension of
V-JEPA 2~\citep{assran2025vjepa2}, predicts future states in latent space,
applying representation prediction to conditional state evolution modeling.
These studies demonstrate different uses of representation prediction in
context modeling and future-state learning. Inspired by these studies, we
use representations of future analysis fields as auxiliary supervision for
concentration forecasting, exploring the integration of representation
prediction with concentration field generation.


\section{Task Definition}
\label{sec:preliminaries}
\label{sec:task_definition}
We study joint forecasting of regional surface PM$_{2.5}$ concentrations and
meteorological fields conditioned on multiple sources of historical information.
Let the study region be an $H\times W$ grid and $t$ the forecast initialization
time. The historical window is
$\mathcal T_t=\{t-(L-1)\Delta,\ldots,t\}$, comprising $L$ frames at intervals
of $\Delta$. Inputs include PM$_{2.5}$ model backgrounds $B$, meteorological
fields $M$, ground observations $O$ and their validity masks $m$, static
geographical information $S$, and time encodings $E$. The model learns the map
\begin{equation}
  \begin{aligned}
    (\widehat C_{t+\Delta},\widehat M_{t+\Delta})
      =F_\theta\bigl(&B_{\mathcal T_t},M_{\mathcal T_t},O_{\mathcal T_t},m_{\mathcal T_t},S,E_{\mathcal T_t}\bigr),
  \end{aligned}
  \label{eq:forecast_task}
\end{equation}
where $\widehat C_{t+\Delta}$ is the future surface PM$_{2.5}$ concentration
field and $\widehat M_{t+\Delta}$ contains ten meteorological variables at the
same target time. We use $L=5$ and $\Delta=3\,\mathrm{h}$ to jointly predict
the next concentration and meteorological fields from five frames spanning
the preceding 12 hours through initialization.


\section{Methods}
\label{sec:framework}

\subsection{Overall Framework}
\label{sec:multisource_future_state}

\begin{figure*}[!t]
  \centering
  \includegraphics[width=0.96\textwidth]{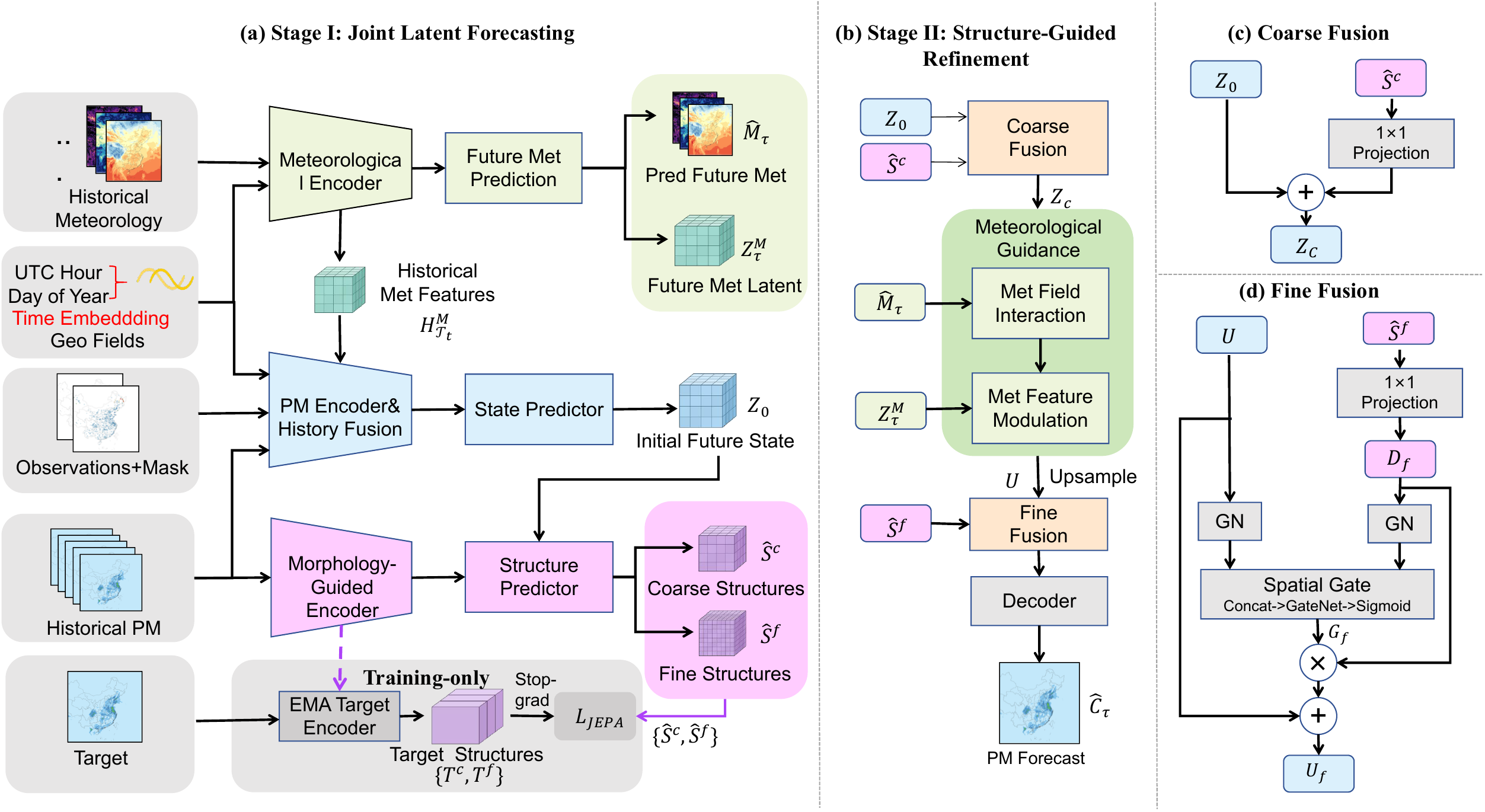}
  \caption{(a) Stage I predicts future latent states, meteorology, and
  multi-scale structures from historical inputs and geographic and temporal
  context. An EMA target encoder provides training-only structure supervision
  from future PM$_{2.5}$ analysis fields.
  (b) Stage II uses the predicted structures and meteorology to refine the
  latent state and decode the PM$_{2.5}$ forecast.
  (c) Coarse fusion projects the predicted coarse structures and adds them
  to the initial latent state.
  (d) Fine fusion injects projected fine structures through a spatial gate
  computed from normalized main-branch and structure features; residual
  injection retains their original magnitudes.}
  \label{fig:overall_framework}
\end{figure*}

As shown in \cref{fig:overall_framework}, AeroJEPA follows a two-stage
forecasting framework, with $\tau=t+\Delta$ denoting the target time.
In Stage I, the meteorological branch encodes historical meteorology and
geographic and temporal information to extract historical meteorological
features and predict future meteorological fields $\widehat M_\tau$ and
latent features $Z^M_\tau$. The PM$_{2.5}$ branch fuses historical model
backgrounds, observations and their masks, context, and historical
meteorological features, then uses a state predictor to generate an initial
future latent state $Z_0$. A parallel structure branch encodes historical
PM$_{2.5}$ fields and predicts coarse and fine structures
$\widehat S^c,\widehat S^f$ conditioned on $Z_0$. Stage II fuses the coarse
structures into $Z_0$, refines the state through interaction with the
predicted meteorological fields and modulation by the latent features,
and then fuses the fine structures and decodes the final concentration
field $\widehat C_\tau$. During training, an EMA target encoder extracts
target structures from future analysis fields to provide auxiliary
supervision for the predicted structures.


\subsection{Stage I: Joint Latent Forecasting}
\label{sec:stage_one}

At each historical time step, PM$_{2.5}$ background features incorporate
geographic and temporal context and undergo FiLM~\citep{perez2018film}
modulation using concurrent meteorological features. Observations,
observation--background differences, and validity masks then generate
gated residual updates, with an observation-support mask restricting their
spatial extent. Historical temporal modeling uses gated fusion of
PM-specific attention and attention recomputed on PM features using
meteorological Q/K projections. Finally, historical feature differences are fused using
the latest frame as an anchor, and an alignment projection and residual
predictor generate the initial future latent state $Z_0$, which conditions
future structure prediction and initializes Stage II refinement.

\begin{figure}[t]
  \centering
  \includegraphics[width=\linewidth]{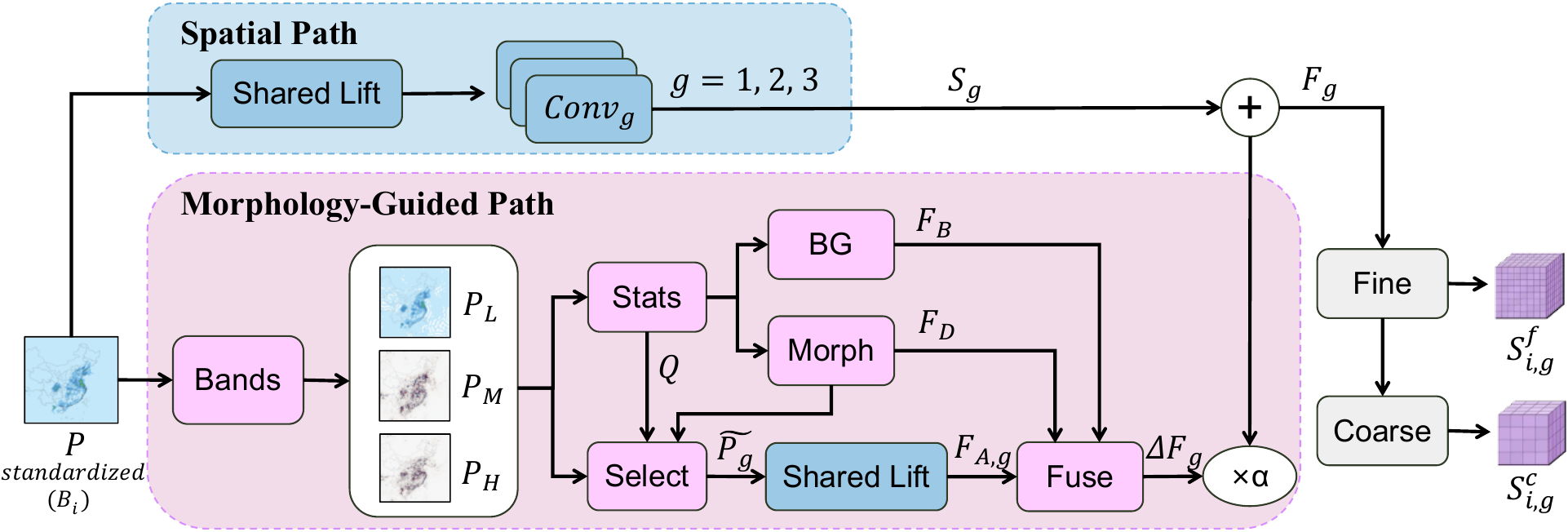}
  \caption{Architecture of the morphology-guided structure encoder. A spatial
  path processes the full PM$_{2.5}$ field, while a morphology-guided path
  combines fixed frequency decomposition, background and morphology encoding,
  and adaptive frequency selection. The two paths are combined through a
  scaled residual update before fine- and coarse-scale structure encoding.}
  \label{fig:structure_encoder}
\end{figure}

As shown in \cref{fig:structure_encoder}, the structure encoder constructs
multi-scale structural representations through a full-field spatial path
and a morphology-guided path. For a standardized historical model-background
PM$_{2.5}$ field $P$, the spatial path uses shared lifting and three
independent convolutional branches to produce group features $S_g$,
$g=1,2,3$. The morphology-guided path filters the two-dimensional Fourier
spectrum of $P$ using fixed smooth radial windows, then applies inverse
transforms to obtain low-, mid-, and high-frequency components $P_L,P_M,P_H$.
The cutoff frequencies are $1/16$ and $1/4$ cycles/pixel, and the windows
sum to one, yielding $P_L+P_M+P_H=P$. These components are adaptively mixed
according to local morphology to enhance the spatial representation
through a residual connection. We construct two smoothed backgrounds, $B_c=P_L$ and
$B_l=P_L+P_M$, and define signed deviations $D_s=P-B_s$, $s\in\{c,l\}$.
To describe both local variation intensity and relative morphology,
we compute local root-mean-square magnitudes and normalized deviations:
\begin{equation}
  \begin{aligned}
    R_s&=\sqrt{\operatorname{AvgPool}_{k_s\times k_s}(D_s^2)+\epsilon^2},\\
    N_s&=D_s/R_s,\qquad s\in\{c,l\}.
  \end{aligned}
  \label{eq:local_morphology_normalization}
\end{equation}

The quantity $R_s$ measures local variation magnitude, while $N_s$ describes
morphology relative to that magnitude; $\epsilon$ is a stabilizing term.
We use $k_c=17$, $k_l=5$, and $\epsilon=0.05$, with $\epsilon$ expressed in standardized PM$_{2.5}$ units.
A background encoder maps $[B_c,B_l]$ to $F_B$, while a morphology encoder
maps $[D_c,D_l,N_c,N_l,\allowbreak\log(1+R_c),\allowbreak\log(1+R_l)]$
to $F_D$. We further construct a local descriptor
$Q=[B_c,B_l,N_c,N_l,\allowbreak\log(1+R_c),\allowbreak\log(1+R_l)]$,
which combines with the morphology features to generate spatially varying
band weights:
\begingroup
\setlength{\abovedisplayskip}{6pt plus 2pt minus 2pt}
\setlength{\belowdisplayskip}{6pt plus 2pt minus 2pt}
\setlength{\abovedisplayshortskip}{3pt plus 1pt}
\setlength{\belowdisplayshortskip}{3pt plus 1pt}
\begin{equation}
  \begin{gathered}
    A=\tanh\!\left(\Phi_A(Q)+\Pi_D(F_D)\right),\\
    \widetilde P_g=\sum\nolimits_{b\in\{L,M,H\}}A_{g,b}\odot P_b,\\
    F_{A,g}=\operatorname{Lift}(\widetilde P_g).
  \end{gathered}
  \label{eq:morphology_guided_band_mixing}
\end{equation}
\endgroup

Here, $\Phi_A$ is a local selector, $\Pi_D$ is a learnable projection,
and $A_{g,b}$ is a spatially varying weight for band $b$ in group $g$ that
can take positive or negative values. Each group first mixes the three
bands and then applies shared linear lifting to obtain $F_{A,g}$.
Finally, the background, morphology, and mixed-band features are
concatenated and projected to produce an enhancement $\Delta F_g$,
which is added to the spatial path with a learnable coefficient $\alpha$:
\begin{equation}
  \begin{aligned}
    \Delta F_g&=\Pi\!\left([F_B,F_D,F_{A,g}]\right),\\
    F_g&=S_g+\alpha\,\Delta F_g.
  \end{aligned}
  \label{eq:morphology_guided_structure_encoding}
\end{equation}

The notation $[\cdot]$ denotes channel concatenation, and $\Pi$ is a shared
projection. A shared local encoder maps the enhanced features $F_g$
successively to fine structures $S_{i,g}^{f}$ at $H/2\times W/2$ resolution
and coarse structures $S_{i,g}^{c}$ at $H/4\times W/4$ resolution,
where $i$ indexes historical time steps. The three groups are adaptive
mixtures and do not correspond to fixed frequency bands.


The coarse and fine structures encoded at each historical time step are
temporally fused within each group to obtain $\bar S_g^c,\bar S_g^f$.
Conditioned on the initial future latent state $Z_0$, we first predict
future coarse structures and then upsample them to guide fine-structure
prediction:
\begin{equation}
  \begin{aligned}
    \widehat S_{\tau,g}^{c}
      &=\bar S_g^c+\Psi_c([\bar S_g^c,Z_0]),\\
    \widehat S_{\tau,g}^{f}
      &=\bar S_g^f+\Psi_f\!\left(
        [\bar S_g^f,\operatorname{Up}(\widehat S_{\tau,g}^{c})]
      \right).
  \end{aligned}
  \label{eq:future_multiscale_structure_prediction}
\end{equation}

The predictors $\Psi_c,\Psi_f$ share parameters across the three groups.
Predictions at each scale are concatenated along channels to form
$\widehat S^c,\widehat S^f$ for Stage II structure fusion.

\subsection{Stage II: Structure-Guided Refinement}
\label{sec:stage_two}

\begin{figure}[t]
  \centering
  \includegraphics[width=0.80\linewidth]{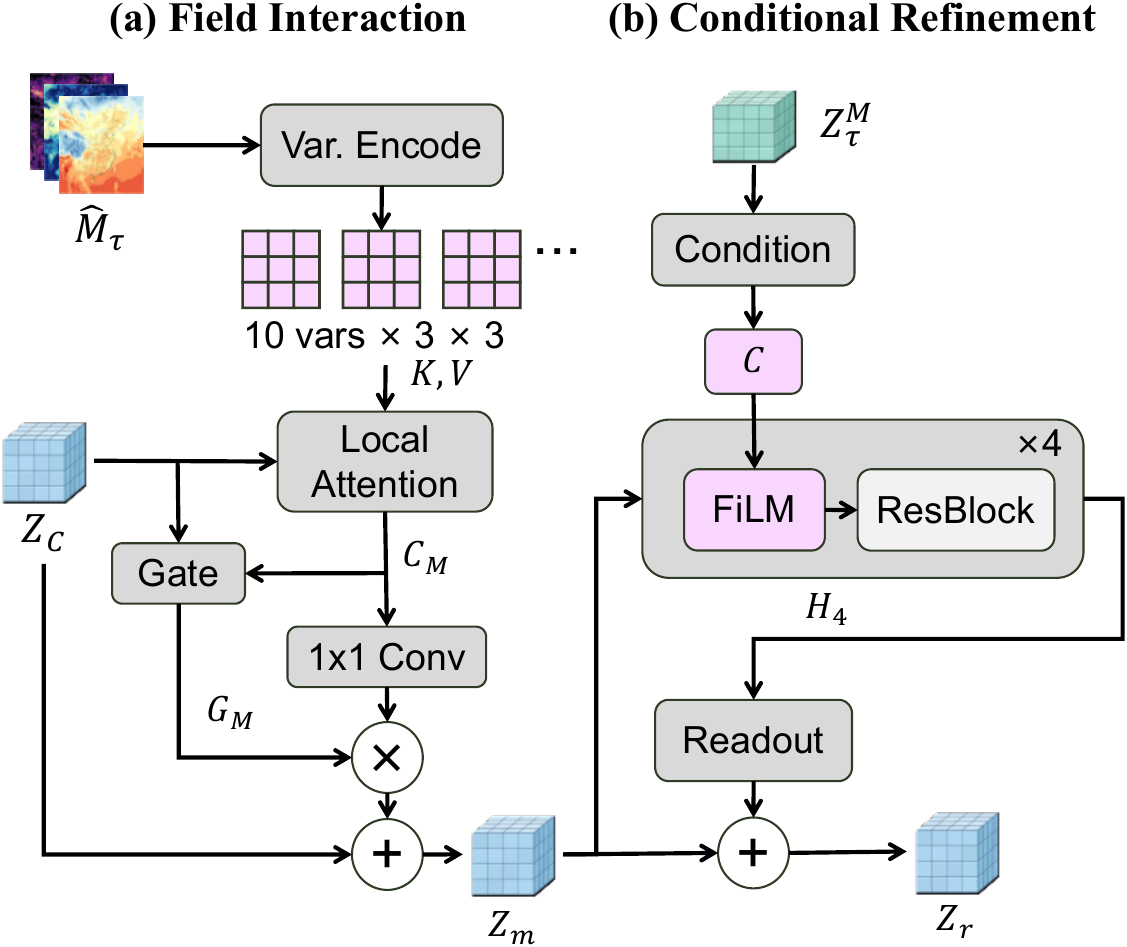}
  \caption{(a) Field interaction. The PM$_{2.5}$ state $Z_c$ attends
  jointly to ten predicted meteorological variables and their $3\times3$
  feature neighborhoods. Gated residual injection produces $Z_m$.
  (b) Conditional refinement. Predicted meteorological latent features
  $Z^M_\tau$ condition four FiLM-modulated residual blocks, followed by
  readout and a residual update to $Z_m$ to produce the refined state $Z_r$.}
  \label{fig:meteorological_modules}
\end{figure}

As shown in \cref{fig:overall_framework}(c), the predicted coarse
structures are projected and added to the initial future latent state,
yielding $Z_c=Z_0+\Pi_c(\widehat S^c)$. Next, as illustrated in
\cref{fig:meteorological_modules}(a), the predicted meteorological fields
$\widehat M_\tau$ are aligned to the state resolution, and the ten
variables are separately encoded with identity embeddings to obtain $T_v$.
Queries are generated from $Z_c(x)$, while keys and values are obtained
from each variable's local $3\times3$ feature neighborhood. Information
is jointly selected across variables and neighboring positions, and the
state is updated through a gated residual connection:
\begin{equation}
  \begin{aligned}
    a_{v,\delta}(x)
      &=\operatorname{softmax}_{v,\delta}\!\left(
        \frac{q_x^\top k_{v,\delta}(x)}{\sqrt d}+b_{v,\delta}\right),\\
    C_M(x)&=\sum_{v=1}^{10}\sum_{\delta\in\mathcal N}
      a_{v,\delta}(x)\,\mathcal V(T_v(x+\delta)),\\
    Z_m&=Z_c+G_M(Z_c,C_M)\odot\Pi_M(C_M),
  \end{aligned}
  \label{eq:future_met_interaction}
\end{equation}
where $\mathcal N$ is the local feature neighborhood, $d$ is the key
dimension, and $b_{v,\delta}$ denotes variable and relative-position
biases. The gate $G_M$ controls injection strength using the current
state and aggregated meteorological features. Joint normalization over
variables and neighboring positions allows each location to select
relevant meteorological information.

The meteorological latent features are further encoded into a condition
$C=\Phi_M(Z^M_\tau)$, which generates FiLM~\citep{perez2018film} scaling and
shifting parameters for each of four successive residual units.
Starting from $H_0=Z_m$, conditional refinement and the final residual
update are given by
\begin{equation}
  \begin{aligned}
    H_l&=\operatorname{RB}_l\!\left(
      (1+\gamma_l(C))\odot H_{l-1}+\beta_l(C)\right),\\
    Z_r&=Z_m+\operatorname{Readout}(H_4),\qquad l=1,\ldots,4.
  \end{aligned}
  \label{eq:future_met_conditional_refinement}
\end{equation}
The refined state $Z_r$ is upsampled and fused with the predicted fine
structures through the gate in \cref{fig:overall_framework}(d), then
decoded into the final concentration field $\widehat C_\tau$.


\subsection{Training and Inference}
\label{sec:training_inference}

To supervise future structure prediction, a target structure encoder extracts
coarse- and fine-scale representations from analysis fields at the target time
during training. Let $C_\tau$ denote the future analysis field and
$\mathcal E_{\bar\phi}$ the target structure encoder. The target structures
and encoder parameter update are given by
\begin{equation}
  (T^c,T^f)=\operatorname{sg}\!\left(\mathcal E_{\bar\phi}(C_\tau)\right),
  \quad \bar\phi\leftarrow\mu\bar\phi+(1-\mu)\phi,
  \label{eq:structure_targets_ema}
\end{equation}
where $\operatorname{sg}$ denotes the stop-gradient operation, $\phi$
contains the online structure encoder parameters, and $\mu$ is the
exponential moving average coefficient. The predicted structures
$\widehat S^c,\widehat S^f$ are matched to the target representations at
the corresponding scales, providing additional structural supervision
alongside concentration regression.


We jointly train both stages end to end using concentration prediction,
meteorological forecasting, and future-structure matching objectives:
\begin{equation}
  \mathcal L=\mathcal L_{\mathrm{PM}}
    +\lambda_{\mathrm{met}}\mathcal L_{\mathrm{met}}
    +\lambda_{\mathrm{JEPA}}\mathcal L_{\mathrm{JEPA}}.
  \label{eq:joint_training_loss}
\end{equation}
The losses $\mathcal L_{\mathrm{PM}}$ and $\mathcal L_{\mathrm{met}}$ are the
mean squared errors in standardized space between concentration predictions
and future analysis fields, and between predicted and target meteorological
fields, respectively. The loss $\mathcal L_{\mathrm{JEPA}}$ matches predicted
structures to stop-gradient targets $T^c,T^f$: feature mean squared errors
are computed after channel-wise layer normalization at each spatial
location, separately for the three groups and two scales, and averaged
with equal weights. Each term equally combines full-domain and historical
observation-neighborhood errors, falling back to the full-domain error when
no valid observation neighborhood is available. The union of the five
historical observation-validity masks is dilated using $9\times9$ max
pooling on the original grid and mapped to each feature scale by adaptive
max pooling.
Loss weights and training
hyperparameters are specified in \cref{sec:experimental_setting}.


During training, the EMA target encoder generates target structures from
future analysis fields, with its parameters updated by an exponential
moving average of the online structure encoder parameters. Inference uses
only information available at and before initialization to perform future
prediction, conditional refinement, and decoding, requiring neither the
target encoder nor ground-truth data at the target time.


\section{Experiments}
\label{sec:experiments}

\subsection{Experimental Setting}
\label{sec:experimental_setting}
\noindent\textbf{Dataset and splits.}
We construct a regional PM$_{2.5}$ forecasting dataset from outputs of the
modified WRF-Chem model developed by Li and colleagues, using the configuration
described by Liu et al.~\citep{liu2019organiccoating,liu2020nitratepathways},
and ground-based monitoring data,
covering 1 January to 8 May 2022. The study region covers a large part of
China on a $400\times400$ grid at 12 km horizontal resolution. Samples are
split chronologically into training, validation, and test sets of 1,711,
389, and 388 samples, respectively, with historical input windows and
prediction targets confined to their respective splits. Observations from
1,597 stations are used as historical inputs and assimilated to construct
labels, while 399 held-out stations are used only for independent
evaluation and excluded from model inputs and label assimilation.
WRF-Chem configurations, assimilation methods, and data preprocessing
details are provided in the supplementary material.


\noindent\textbf{Compared methods.}
We compare AeroJEPA with WRF numerical forecasts and six deep learning
baselines: AFNO~\citep{guibas2022afno}, Earthformer~\citep{gao2022earthformer},
PFGNet~\citep{cai2026pfgnet}, SimVP~\citep{gao2022simvp,tan2025simvpv2},
Swin-UNet~\citep{cao2023swinunet}, and MCVD~\citep{voleti2022mcvd}.
All methods predict surface PM$_{2.5}$ concentrations 3 h after
initialization. AeroJEPA and all six deep learning baselines use the same
historical inputs: five frames spanning 12 h at 3 h intervals, comprising
PM$_{2.5}$ model backgrounds, ten meteorological variables, sparse
PM$_{2.5}$ observations and their validity masks, together with time
encodings and static information. All deep learning models are adapted
to jointly predict PM$_{2.5}$ concentrations and meteorological fields,
using identical data splits, training-set normalization, and supervision
targets. Baseline configurations and task-specific adaptations are
detailed in the supplementary material.


\noindent\textbf{Evaluation metrics.}
Gridded evaluation against target-time 3D-Var analysis fields uses root mean
squared error (RMSE), mean absolute error (MAE), mean bias (BIAS), spatial
correlation (CORR), and structural similarity (SSIM)~\citep{wang2004ssim}.
Independent-station evaluation uses RMSE, MAE, BIAS, and Pearson correlation
$r$. All methods use the same data splits and valid evaluation pairs.


\noindent\textbf{Training settings.}
All deep learning methods are trained with AdamW~\citep{loshchilov2019adamw},
using a learning rate and weight decay of $10^{-4}$ and a batch size of 1.
Training runs for at most 100 epochs, with checkpoint selection and early
stopping based on validation PM$_{2.5}$ MSE and a patience of 15 epochs.
The five deterministic baselines minimize the sum of PM$_{2.5}$ MSE and
mean meteorological MSE in normalized space. AeroJEPA additionally applies
JEPA structure matching with a weight of 0.2; its meteorological loss
weight and target-encoder EMA momentum are 1 and 0.996, respectively.
MCVD uses a diffusion training objective, with its ensemble mean used as
the point forecast for evaluation. Full training, sampling, and
postprocessing settings are provided in the supplementary material.


\subsection{Experimental Results}
\label{sec:experimental_results}
\begin{table*}[!t]
  \centering
  \caption{Single-step (+3 h) PM$_{2.5}$ forecasting results evaluated against
  gridded analysis fields and independent-station observations on the test set. Station scores use 129,159 common valid pairs
  from 343 stations with observations among 399 fixed independent stations.
  WRF uses the original forecast at the target time. Analysis is a reference
  excluded from rankings.
  RMSE, MAE, and BIAS are in $\mu\mathrm{g}\,\mathrm{m}^{-3}$.
  BIAS values closer to zero are preferred. Best forecasting scores are bold.}
  \label{tab:pm25_results}
  \label{tab:pm25_grid_results}
  \label{tab:pm25_station_results}
  \scriptsize
  \setlength{\tabcolsep}{3pt}
  \renewcommand{\arraystretch}{1.12}
  \begin{tabular*}{\textwidth}{@{\extracolsep{\fill}}lrrrrrrrrr@{}}
    \toprule
    & \multicolumn{5}{c}{Grid evaluation} & \multicolumn{4}{c}{Station evaluation} \\
    \cmidrule(lr){2-6}\cmidrule(l){7-10}
    Method
      & RMSE$\downarrow$ & MAE$\downarrow$ & BIAS & CORR$\uparrow$ & SSIM$\uparrow$
      & RMSE$\downarrow$ & MAE$\downarrow$ & BIAS & $r\uparrow$ \\
    \midrule
    Analysis (reference) & -- & -- & -- & -- & -- & 11.9270 & 7.4488 & +0.0562 & 0.8486 \\
    \midrule
    WRF~\cite{grell2005wrfchem} & 5.6631 & \textbf{1.3139} & -0.6717 & 0.8666 & 0.9678 & 26.8484 & 16.7198 & -5.7767 & 0.2315 \\
    AFNO~\cite{guibas2022afno} & 4.8355 & 1.9750 & -0.5545 & 0.9054 & 0.9644 & 18.7309 & 10.3415 & -4.5823 & 0.5804 \\
    Earthformer~\cite{gao2022earthformer} & 5.6275 & 2.9433 & +1.4166 & 0.8740 & 0.9173 & 20.5593 & 12.0265 & -1.1461 & 0.4723 \\
    PFGNet~\cite{cai2026pfgnet} & 4.5123 & 2.2720 & +0.6142 & 0.9155 & 0.9459 & 16.9445 & 9.5140 & -2.7953 & 0.6622 \\
    SimVP~\cite{gao2022simvp,tan2025simvpv2} & 5.7908 & 2.8194 & +0.6774 & 0.8606 & 0.9212 & 20.6911 & 12.3532 & -0.2242 & 0.4655 \\
    Swin-UNet~\cite{cao2023swinunet} & 5.1592 & 2.1971 & -0.4580 & 0.8841 & 0.9520 & 19.8591 & 11.5513 & -3.1510 & 0.5225 \\
    MCVD~\cite{voleti2022mcvd} & 9.8629 & 6.4530 & +2.4620 & 0.5960 & 0.7282 & 24.7283 & 16.3413 & -12.0242 & 0.3678 \\
    \midrule
    \textbf{AeroJEPA(ours)} & \textbf{3.7347} & 1.3727 & \textbf{+0.1157} & \textbf{0.9410} & \textbf{0.9801} & \textbf{15.6055} & \textbf{8.9285} & \textbf{+0.0521} & \textbf{0.7145} \\
    \bottomrule
  \end{tabular*}

  \par\medskip
  \normalsize
  \begingroup
    \def\pmResultsInline{}
    \noindent
    \begin{minipage}[t]{0.48\textwidth}
      \vspace{0pt}
\ifdefined\pmResultsInline
\captionsetup{type=figure}
\else
\begin{figure}[htbp]
\fi
  \centering
  \includegraphics[width=\linewidth]{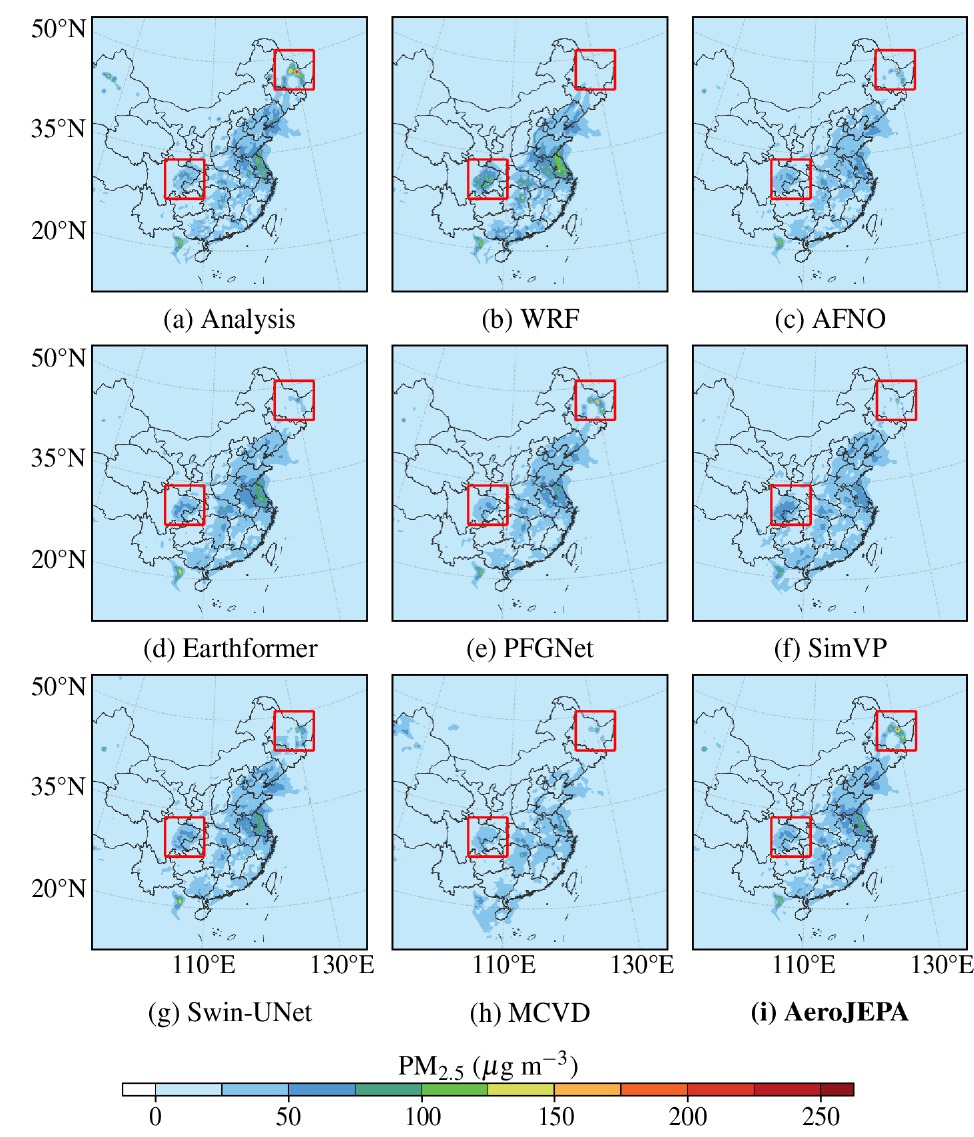}
  \caption{PM$_{2.5}$ fields for a +3 h forecast initialized at 15:00 UTC
  on 19 April 2022. Analysis and all forecasts correspond to $t+3$ h and
  share the same color scale. Units: $\mu\mathrm{g}\,\mathrm{m}^{-3}$.}
  \label{fig:pm25_forecast_comparison}
\ifdefined\pmResultsInline\else
\end{figure}
\fi

    \end{minipage}\hfill
    \begin{minipage}[t]{0.48\textwidth}
      \vspace{0pt}
\ifdefined\pmResultsInline
\captionsetup{type=figure}
\else
\begin{figure}[t]
\fi
  \centering
  \includegraphics[width=\linewidth]{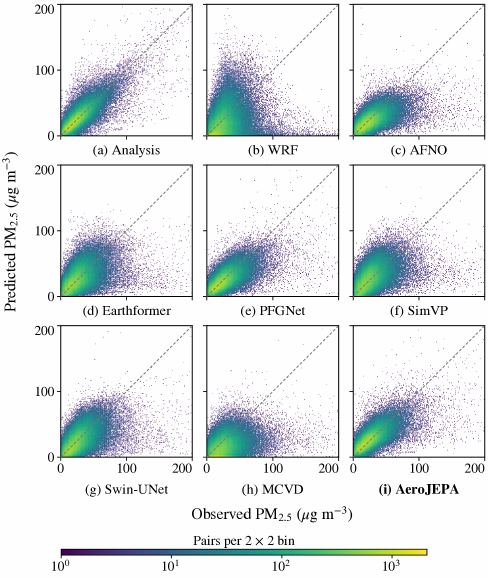}
  \caption{Test-set density scatter plots against independent-station
  observations. Analysis is a target-time reference. Dashed lines indicate
  $y=x$; the shared logarithmic color scale shows pair counts in $2\times2$
  concentration bins. Both axes are in $\mu\mathrm{g}\,\mathrm{m}^{-3}$.}
  \label{fig:pm25_station_scatter_comparison}
\ifdefined\pmResultsInline\else
\end{figure}
\fi

    \end{minipage}
  \endgroup
\end{table*}

\noindent\textbf{Gridded Forecasting Results.}
As shown in \cref{tab:pm25_results}, AeroJEPA achieves the lowest RMSE,
the BIAS closest to zero, and the highest CORR and SSIM in gridded
evaluation. Its RMSE is $3.7347\,\mu\mathrm{g}\,\mathrm{m}^{-3}$,
reductions of 17.23\% and 34.05\% relative to PFGNet, the deep learning
baseline with the lowest RMSE, and WRF, respectively. Its MAE is lower
than those of all deep learning baselines but slightly higher than WRF's.
The +3 h case in \cref{fig:pm25_forecast_comparison} shows that AeroJEPA
preserves the main concentration distribution over central and eastern
China and, like PFGNet, reproduces the local high-concentration region
in northeastern China, which some baselines substantially underestimate.
Some deviations in high-concentration predictions remain despite the
improved overall structural agreement with the analysis targets.


\noindent\textbf{Independent-station evaluation.}
At independent stations excluded from historical inputs and label
assimilation, AeroJEPA also achieves the lowest RMSE and MAE, the highest
correlation, and the BIAS closest to zero among all forecasting methods.
RMSE and MAE are $15.6055$ and $8.9285\,\mu\mathrm{g}\,\mathrm{m}^{-3}$,
respectively, reductions of 7.90\% and 6.15\% relative to PFGNet, the deep
learning baseline with the lowest independent-station errors.
These results show that the gridded improvements also translate to
predictions against independent ground observations.
In \cref{fig:pm25_station_scatter_comparison}, AeroJEPA's main point cloud
is more concentrated around the agreement line than those of WRF and the
other deep learning methods, although high observed concentrations
remain underestimated.


\subsection{Ablation Study}
\label{sec:ablation_study}

\begin{table}[!t]
  \centering
  \caption{Ablation results for single-step PM$_{2.5}$ forecasting against
  target-time analysis fields. MSE is in $(\mu\mathrm{g}\,\mathrm{m}^{-3})^2$;
  MAE is in $\mu\mathrm{g}\,\mathrm{m}^{-3}$.
  CFF: coarse- and fine-scale fusion; CMG: cross-stage meteorological
  guidance; MGP: morphology-guided path.
  Best results are shown in bold.}
  \label{tab:pm25_ablation_results}
  \footnotesize
  \setlength{\tabcolsep}{1.5pt}
  \renewcommand{\arraystretch}{1.15}
  \begin{tabular*}{\linewidth}{@{\extracolsep{\fill}}lrrrr@{}}
    \toprule
    Model & MSE$\downarrow$ & MAE$\downarrow$ & CORR$\uparrow$ & SSIM$\uparrow$ \\
    \midrule
    \textbf{AeroJEPA} & \textbf{13.94819} & \textbf{1.37274} & \textbf{0.94096} & \textbf{0.98007} \\
    w/o JEPA & 14.52568 & 1.39725 & 0.93928 & 0.97913 \\
    w/o JEPA (MS Sup.) & 14.38698 & 1.44812 & 0.93999 & 0.97841 \\ 
    w/o CFF & 14.14177 & 1.42873 & 0.94014 & 0.97898 \\
    w/o CMG & 14.16526 & 1.42172 & 0.94034 & 0.97900 \\
    w/o MGP & 14.34494 & 1.37563 & 0.94026 & 0.97999 \\
    \bottomrule
  \end{tabular*}
\end{table}

To assess each component, we separately disable future-structure matching
supervision (w/o JEPA), coarse- and fine-scale fusion (w/o CFF), and
cross-stage meteorological guidance (w/o CMG), and replace the
morphology-guided enhancement front end with standard spatial convolutions
(w/o MGP). We also replace JEPA with multi-scale supervision, denoted as
\textit{w/o JEPA (MS Sup.)}, using auxiliary readouts of the predicted coarse
and fine structures for PM$_{2.5}$ reconstruction and Sobel-gradient
matching, while retaining structure prediction and both fusion paths.
The auxiliary loss weight is selected by validation PM$_{2.5}$ MSE.
All variants use the same data splits, optimization settings, and
evaluation protocol as the complete model.


As shown in \cref{tab:pm25_ablation_results}, the complete model performs
best on all four metrics. With structure prediction and fusion retained,
removing JEPA supervision increases MSE by 4.14\%, the largest increase
among the variants, indicating benefits beyond final concentration
regression. The \textit{w/o JEPA (MS Sup.)} variant increases MSE and MAE
by 3.15\% and 5.49\%, respectively, and lowers CORR and SSIM, favoring
representation matching over this fixed structural supervision objective.
With JEPA retained, disabling coarse- and fine-scale fusion increases
MAE by 4.08\% and reduces SSIM, supporting the direct use of predicted
structures in state updates and decoding.
Removing cross-stage meteorological guidance increases MSE and MAE by
1.56\% and 3.57\%, respectively. Since the meteorological forecasting
task is retained, this supports using its predictions as refinement
conditions. Replacing the morphology-guided enhancement front end with
standard spatial convolutions increases MSE by 2.84\% but MAE by only
0.21\%, indicating benefits mainly in MSE.


\subsection{Multi-Step Forecasting Case Study}
\label{sec:multistep_forecasting_comparison}

\begin{figure}[t]
  \centering
  \includegraphics[width=\linewidth]{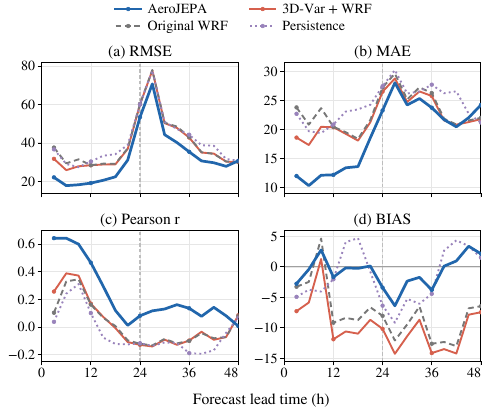}
  \caption{Independent-station PM$_{2.5}$ metrics across lead times in a
  48-hour multi-step forecasting experiment. Vertical dashed lines mark
  the 24 h lead time. RMSE, MAE, and BIAS are in
  $\mu\mathrm{g}\,\mathrm{m}^{-3}$.}
  \label{fig:multistep_forecasting_comparison}
\end{figure}

\noindent\textbf{Forecasting protocol.}
In the single-step experiments above, AeroJEPA uses historical ground
observations, whereas the original WRF forecasts do not incorporate these
observational constraints. Their performance differences therefore also
reflect the benefits of using observations and cannot be attributed
entirely to the forecasting methods themselves. To examine the single-step
model under recursive feedback, we conduct a 48 h forecasting case study
initialized at 15:00 UTC on 19 April 2022. We compare AeroJEPA with
Original WRF, Persistence, and 3D-Var + WRF, which assimilates observations
only at initialization and then integrates continuously. Starting from
five historical frames, AeroJEPA
predicts PM$_{2.5}$ and meteorological fields 3 h ahead and feeds its
predictions back into the sliding window for 16 consecutive steps.
Observation inputs for newly added frames are set to zero with invalid
masks, while historical observations gradually leave the window; no new
observations or ground-truth future meteorology enter the multi-step forecasting procedure.
All methods are paired with independent-station observations at the same
valid times to compute lead-time RMSE, MAE, Pearson correlation $r$, and
BIAS.


\noindent\textbf{Forecast performance.}
As shown in \cref{fig:multistep_forecasting_comparison}, in this case,
the RMSE and MAE advantages of 3D-Var + WRF over Original WRF narrow
with increasing lead time, suggesting a diminishing benefit from initial
assimilation. During the first 24 h, AeroJEPA generally achieves lower
RMSE and MAE and higher correlation than the numerical baselines.
Errors rise markedly for all methods around +24 to +27 h. AeroJEPA
retains lower RMSE at most lead times, but its MAE advantage weakens
and correlation declines during the second day. Some metrics approach
baseline levels near the end, indicating limitations at longer lead times.
Forecast-window and full-period results are provided in the supplementary
material.


\section{Conclusion}
\label{sec:conclusion}
We presented AeroJEPA, a two-stage framework for regional surface PM$_{2.5}$
forecasting that integrates historical model backgrounds, ground observations,
and meteorological information. The first stage predicts an initial future
latent state, multi-scale spatial structures, and meteorological conditions.
The second stage combines coarse- and fine-scale structure fusion with
cross-stage meteorological guidance to refine the future latent state and
generate the concentration field. Morphology-guided structure encoding and
JEPA-style matching with an EMA target encoder enable learning of predictive
spatial representations alongside concentration regression.
At +3 h, evaluations against gridded analysis fields and observations
from held-out stations, together with component ablations, support the
effectiveness of the framework. A 48 h recursive
forecasting case study with a single initialization also shows lower RMSE
and MAE than 3D-Var + WRF at most lead times, illustrating its potential
for recursive forecasting under repeated prediction feedback without new
observations. These findings support combining predictive spatial representations
with meteorological conditioning. The reduced
advantage at longer lead times motivates further investigation of error
accumulation and stability under recursive forecasting.

%

\FloatBarrier
\section*{AI Disclosure Statement}
Generative AI tools were used to assist with methodological discussion, code development and debugging, and literature search. They were also used to improve the clarity, grammar, and organization of the manuscript. All AI-assisted methodological suggestions were reviewed and validated by the authors, all generated or modified code was tested by the authors, and cited literature was checked against the original sources. Generative AI tools were not used to generate experimental data or replace the authors' scientific judgment. The authors take full responsibility for the final content of this work, including all text, claims, results, and artifacts produced with the aid of generative AI.

\bibliography{main}
\bibliographystyle{icml2027}

\clearpage
\appendix

\twocolumn[{
  \centering
  {\Large\bfseries Appendix\par}
  \vspace{1em}
}]

\section{Dataset Details}
\label{sec:supp_dataset}

\noindent\textbf{Data sources.}
The dataset covers 1 January--8 May 2022. Background fields are obtained
from WRF-Chem simulations without data assimilation.
All fields retain the native $400\times400$ grid at a horizontal
resolution of 12 km. The parameterization schemes are listed in
\cref{tab:supp_wrf_config}.
Of the 1,996 valid ground monitoring stations within the study region,
1,597 are used for assimilation and historical observation inputs,
and 399 are held out for independent evaluation.
The station split remains fixed across all dates and pollutants.
Independent stations are excluded from both assimilation and model
inputs (\cref{fig:supp_stations}).


\begin{table}[htbp]
  \centering
  \caption{WRF-Chem parameterization schemes.}
  \label{tab:supp_wrf_config}
  \small
  \begin{tabular}{ll}
    \toprule
    Process & Scheme \\
    \midrule
    Microphysics & Lin \\
    Longwave radiation & RRTM \\
    Shortwave radiation & Goddard \\
    Land surface & Noah \\
    Planetary boundary layer & YSU \\
    Gas-phase chemistry & CBM-Z \\
    Aerosols & MOSAIC\_4bins \\
    \bottomrule
  \end{tabular}
\end{table}

\par\medskip\noindent
\begin{minipage}{\linewidth}
  \captionsetup{type=figure}
  \centering
  \includegraphics[width=0.6\linewidth,trim=0 18bp 0 0,clip]{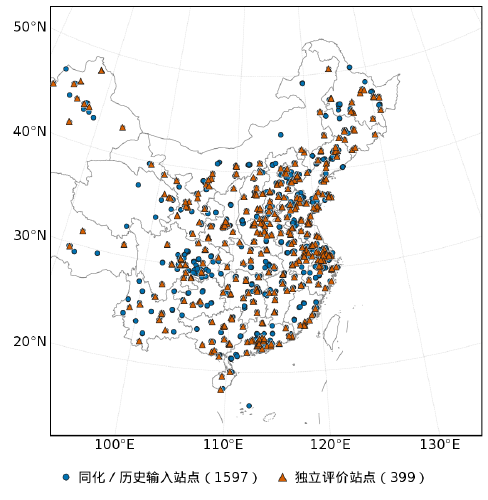}
  \par\smallskip
  {\scriptsize
    \textcolor[RGB]{0,158,230}{$\bullet$} Assimilation/input (1,597)
    \quad
    \textcolor[RGB]{255,127,14}{$\blacktriangle$} Independent (399)}
  \caption{Study region and station distribution. Blue circles denote
  stations used for assimilation and historical inputs; orange triangles
  denote independent evaluation stations.}
  \label{fig:supp_stations}
\end{minipage}
\par\medskip

\noindent\textbf{Analysis targets.}
Analysis fields are generated using a 3D-Var system based on the method
of Li et al.~\citep{li2013aerosol3dvar}.
Observations of SO$_2$, NO$_2$, CO, O$_3$, PM$_{2.5}$, and PM$_{10}$
are assimilated into the three-dimensional WRF-Chem background fields.
The resulting lowest-layer PM$_{2.5}$ fields serve as forecasting targets.
At each target time, an analysis is computed independently using the
original background field and concurrent observations.
Historical model inputs are the original WRF-Chem outputs.

\begin{figure}[t]
  \centering
  \includegraphics[width=\linewidth]{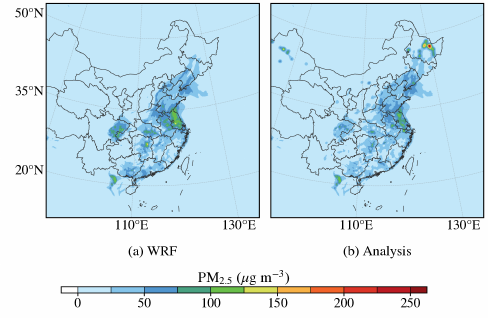}
  \caption{Surface PM$_{2.5}$ fields from (a) WRF and
  (b) Analysis (3D-Var), shown with a shared color scale.
  Units: $\mu\mathrm{g}\,\mathrm{m}^{-3}$.}
  \label{fig:supp_assimilation}
\end{figure}

\Cref{fig:supp_assimilation} compares the WRF background (WRF) with
the 3D-Var analysis (Analysis). Compared with the broad, spatially
connected concentration patterns in WRF, Analysis exhibits more
localized variations and sharper concentration contrasts in this example.


\noindent\textbf{Preprocessing.}
Observation timestamps are converted from Beijing time to UTC.
Missing, non-finite, and negative observations are treated as invalid.
To construct historical observation inputs, valid PM$_{2.5}$ measurements
are mapped to the nearest WRF grid point. Measurements from multiple
stations assigned to the same grid cell are averaged.
Cells without valid observations are filled with zero and marked
invalid in the corresponding mask.
The ten meteorological variables are listed in \cref{tab:supp_meteorology}.
Wind components are rotated from grid-relative to geographic directions,
RH$_2$ is constrained to $[0,100]\%$, and three-hour precipitation is
calculated by differencing accumulated precipitation at times
separated by 3 h.


\begin{table}[!t]
  \centering
  \caption{Meteorological variables.}
  \label{tab:supp_meteorology}
  \footnotesize
  \setlength{\tabcolsep}{4pt}
  \begin{tabular}{lll}
    \toprule
    Variable & Description & Unit \\
    \midrule
    U10, V10 & 10 m wind components & $\mathrm{m\,s^{-1}}$ \\
    T2 & 2 m temperature & K \\
    RH2 & 2 m relative humidity & \% \\
    PBLH & Boundary-layer height & m \\
    UST & Friction velocity & $\mathrm{m\,s^{-1}}$ \\
    HFX & Surface sensible heat flux & $\mathrm{W\,m^{-2}}$ \\
    SWDOWN & Downward shortwave flux & $\mathrm{W\,m^{-2}}$ \\
    PRECIP\_3H & Three-hour precipitation & mm \\
    PSFC & Surface pressure & Pa \\
    \bottomrule
  \end{tabular}
\end{table}

Dynamic fields are standardized per channel using statistics computed
only from the unique historical WRF frames used by the training set,
except RH$_2$, which is divided by 100.
PM$_{2.5}$ backgrounds, valid observations, and analysis targets share
the same training-background mean and standard deviation.
Invalid observation cells are reset to zero after standardization.
Latitude and longitude are scaled to $[-1,1]$, terrain is standardized,
and UTC hour and day of year are encoded using sine and cosine functions.


\noindent\textbf{Sample construction.}
Forecast initialization times advance in one-hour increments.
Each sample uses five historical frames spanning $t-12\,\mathrm{h}$ to
$t$ at 3 h intervals, with PM$_{2.5}$ analysis and meteorological targets
at $t+3\,\mathrm{h}$.
The 2,488 samples are split chronologically into training,
validation, and test sets of 1,711, 389, and 388 samples,
respectively (68.8\%, 15.6\%, and 15.6\%).


\section{Implementation Details}
\label{sec:supp_implementation}

\noindent\textbf{Architecture of AeroJEPA.}
For an input grid of size $H\times W$, the main latent state has
64 channels. PM$_{2.5}$ and meteorological features are fused at
$H/4\times W/4$ resolution.
The structure branch produces fine- and coarse-scale features for each
of three structure groups at $H/2\times W/2$ and $H/4\times W/4$
resolution, respectively. The module configurations are summarized
in \cref{tab:supp_model_config}.

%
\begingroup
\setlength{\intextsep}{6pt}
\begin{table}[htbp]
  \centering
  \caption{AeroJEPA configurations.}
  \label{tab:supp_model_config}
  \scriptsize
  \setlength{\tabcolsep}{3pt}
  \renewcommand{\arraystretch}{0.9}
  \setlength{\aboverulesep}{0.3pt}
  \setlength{\belowrulesep}{0.3pt}
  \begin{tabular}{@{}p{0.28\linewidth}|p{0.67\linewidth}@{}}
    \toprule
    \multicolumn{2}{c}{Historical Encoding} \\
    \midrule
    PM encoder & \mbox{Stride-2 Conv$\times2$ ($1\to64$)} \tabularnewline
    Met encoder & \mbox{3D Conv (base 32, latent 64)} \tabularnewline
    Temporal attention & \mbox{QKV Linear + 4-head Attn + Linear} \tabularnewline
    \bottomrule
  \end{tabular}
  \par
  \begin{tabular}{@{}p{0.28\linewidth}|p{0.67\linewidth}@{}}
    \toprule
    \multicolumn{2}{c}{Structure Encoder} \\
    \midrule
    Spatial path & \mbox{Lift$_{1\to8}$ + $3\times3$ Conv} \tabularnewline
    Morphology encoder & \mbox{Conv$_{3\times3}$ + GELU + Conv$_{1\times1}$} \tabularnewline
    Local encoder & \mbox{PixelUnshuffle(2) + $1\times1$ Conv + ResBlock} \tabularnewline
    \bottomrule
  \end{tabular}
  \par
  \begin{tabular}{@{}p{0.28\linewidth}|p{0.67\linewidth}@{}}
    \toprule
    \multicolumn{2}{c}{Prediction and Refinement} \\
    \midrule
    ResBlock & \mbox{DWConv$_{5\times5}$ + GN + Conv$_{1\times1}$ + GELU + Conv$_{1\times1}$} \tabularnewline
    State predictor & \mbox{Align + ResBlock$\times2$ + Conv$_{1\times1}$} \tabularnewline
    Structure predictors & \mbox{Fusion + ResBlock$\times2$ + Conv$_{1\times1}$} \tabularnewline
    Conditional refiner & \mbox{(FiLM + ResBlock)$\times4$ + Conv$_{1\times1}$} \tabularnewline
    \bottomrule
  \end{tabular}
  \par
  \begin{tabular}{@{}p{0.28\linewidth}|p{0.67\linewidth}@{}}
    \toprule
    \multicolumn{2}{c}{Structure Fusion} \\
    \midrule
    Coarse fusion & \mbox{$1\times1$ Conv ($192\to64$) + Add} \tabularnewline
    Fine fusion & \mbox{$1\times1$ Conv ($96\to64$) + Gate} \tabularnewline
    Fine gate & \mbox{Conv$_{1\times1}$ + SiLU + Conv$_{3\times3}$ + Sigmoid} \tabularnewline
    \bottomrule
  \end{tabular}
\end{table}
\endgroup

All three structure groups use a shared local structure encoder and
shared future-structure predictors.
Coarse and fine features are concatenated along the channel dimension
and projected to 64 channels using $1\times1$ convolutions for state
updates and decoding, respectively.
The output projection of the morphology-guided enhancement branch is
zero-initialized, with the residual scaling coefficient initialized
to 0.1. The fine-structure fusion gate is initialized to approximately 0.1.
Predictions are upsampled bilinearly, denormalized, and clipped
at zero for evaluation and inference.


\makeatletter
\providecommand{\theHALG@line}{}
\renewcommand{\theHALG@line}{\thealgorithm.\arabic{ALG@line}}
\makeatother

\renewcommand{\algorithmicrequire}{\textbf{Input:}}
\renewcommand{\algorithmicensure}{\textbf{Output:}}

\textbf{Training and Inference.}
Both stages are trained jointly with PM$_{2.5}$, meteorological,
and JEPA losses (Algorithm~\ref{alg:framework_training}).
The target encoder remains in evaluation mode and receives no gradients;
its parameters are updated by EMA after each optimizer step.
Gradient norms are clipped to 1.
Inference follows the same prediction path without observation dropout
or the target encoder, followed by denormalization and nonnegative
clipping (Algorithm~\ref{alg:framework_inference}).
Here, $(B,M,O,m)$ denote historical backgrounds, meteorology,
observations, and validity masks; $\mathcal{C}$ collects static
information and time encodings. Equation references in
Algorithms~\ref{alg:framework_training} and~\ref{alg:framework_inference}
refer to the main paper.


\begin{algorithm}[t]
\caption{Training of AeroJEPA}
\label{alg:framework_training}
\footnotesize
\begin{algorithmic}[1]
\REQUIRE Training set $\mathcal{D}$; online parameters $\theta$;
structure encoder parameters $\phi\subset\theta$; EMA momentum $\mu$
\ENSURE Trained parameters $\theta$
\STATE $\bar\phi\gets\phi$; set the online model to training mode
\WHILE{not converged}
    \STATE Sample $(B,M,O,m,\mathcal{C},C_\tau,M_\tau)\sim\mathcal{D}$
    \STATE Apply observation dropout: $\widetilde m\gets\operatorname{ObsDrop}(m)$
    \item[] \textit{Stage I}
    \STATE $(H_M,Z^M_\tau,\widehat M_\tau)\gets\operatorname{MetBranch}_\theta(M,\mathcal{C})$
    \STATE $H_P\gets\operatorname{HistoryEncoder}_\theta(B,O,\widetilde m,\mathcal{C},H_M)$
    \STATE $Z_0\gets\operatorname{FutureStatePredictor}_\theta(H_P)$
    \STATE $S^h\gets\mathcal{E}_\phi(B)$
    \STATE Predict $(\widehat S^c,\widehat S^f)$ using Eq.~(\ref*{eq:future_multiscale_structure_prediction})
    \item[] \textit{Stage II}
    \STATE $Z_c\gets Z_0+\Pi_c(\widehat S^c)$
    \STATE Compute $Z_m$ using Eq.~(\ref*{eq:future_met_interaction})
    \STATE Compute $Z_r$ using Eq.~(\ref*{eq:future_met_conditional_refinement})
    \STATE Decode $\widehat C_\tau$ with fine fusion (main paper, Fig.~\ref*{fig:overall_framework}d)
    \item[] \textit{Loss and parameter updates}
    \STATE Encode targets $(T^c,T^f)$ using Eq.~(\ref*{eq:structure_targets_ema})
    \STATE $\Omega\gets\operatorname{Dilate}_4\!\left(\bigvee_{i=1}^5m_i\right)$
    \STATE Compute prediction losses $\mathcal L_{\mathrm{PM}}$ and $\mathcal L_{\mathrm{met}}$
    \STATE Compute $\mathcal L_{\mathrm{JEPA}}$ with support $\Omega$ (main paper, Sec.~\ref*{sec:training_inference})
    \STATE Compute $\mathcal L$ using Eq.~(\ref*{eq:joint_training_loss})
    \STATE Update $\theta$ using AdamW and clipped $\nabla_\theta\mathcal L$
    \STATE Update $\bar\phi$ by EMA using Eq.~(\ref*{eq:structure_targets_ema})
\ENDWHILE
\end{algorithmic}
\end{algorithm}

\begin{algorithm}[t]
\caption{Inference of AeroJEPA}
\label{alg:framework_inference}
\footnotesize
\begin{algorithmic}[1]
\REQUIRE Historical inputs $(B,M,O,m)$; context $\mathcal{C}$;
trained parameters $\theta$; inverse normalization $\mathcal N_{\mathrm{PM}}^{-1}$
\ENSURE Future PM$_{2.5}$ concentration $\widehat C_{\tau,\mathrm{phys}}$
\STATE Set evaluation mode and disable gradients
\STATE $(H_M,Z^M_\tau,\widehat M_\tau)\gets\operatorname{MetBranch}_\theta(M,\mathcal{C})$
\STATE $H_P\gets\operatorname{HistoryEncoder}_\theta(B,O,m,\mathcal{C},H_M)$
\STATE $Z_0\gets\operatorname{FutureStatePredictor}_\theta(H_P)$
\STATE $S^h\gets\mathcal{E}_\phi(B)$
\STATE Predict $(\widehat S^c,\widehat S^f)$ using Eq.~(\ref*{eq:future_multiscale_structure_prediction})
\STATE $Z_c\gets Z_0+\Pi_c(\widehat S^c)$
\STATE Compute $Z_m$ using Eq.~(\ref*{eq:future_met_interaction})
\STATE Compute $Z_r$ using Eq.~(\ref*{eq:future_met_conditional_refinement})
\STATE Decode $\widehat C_\tau$ with fine fusion (main paper, Fig.~\ref*{fig:overall_framework}d)
\STATE \textbf{return} $\max\!\left(0,\mathcal N_{\mathrm{PM}}^{-1}(\widehat C_\tau)\right)$
\end{algorithmic}
\end{algorithm}

\textbf{Experimental Details.}
All deep learning models were trained separately on a single NVIDIA
GeForce RTX 3090 GPU with 24 GB of memory, using FP32 precision and a
batch size of 1. Our implementation uses PyTorch 2.3.0 with CUDA 11.8.
AFNO, Earthformer, PFGNet, Swin-UNet, and MCVD were adapted from their
official implementations, while SimVP-gSTA was implemented locally
following the architecture used in OpenSTL.
For MCVD, validation and testing used EMA weights, 20-step DDIM
sampling, and four ensemble members per input. The ensemble mean was
denormalized and clipped at zero to obtain the PM$_{2.5}$ point forecast.


\section{Additional Results}
\label{sec:supp_additional_results}

\subsection{Multi-Step Forecasting}
\label{sec:supp_multistep_results}

We report additional forecast-window results for the 48 h multi-step
forecasting experiment initialized at 15:00 UTC on 19 April 2022,
following the protocol described in the main paper.
Metrics for each window are computed from all valid prediction--observation
pairs pooled across independent stations, rather than by averaging
lead-time scores.


\begin{table}[t]
  \centering
  \caption{Independent-station PM$_{2.5}$ results aggregated over forecast
  windows for the 48-hour multi-step forecasting experiment. RMSE, MAE, and BIAS are in
  $\mu\mathrm{g}\,\mathrm{m}^{-3}$; $r$ denotes Pearson correlation.
  BIAS values closer to zero are preferred. $N$ denotes the number of valid
  station--time pairs. Best results within each window are shown in bold.}
  \label{tab:pm25_multistep_results}
  \footnotesize
  \setlength{\tabcolsep}{3pt}
  \renewcommand{\arraystretch}{1.08}
  \begin{tabular*}{\linewidth}{@{\extracolsep{\fill}}lrrrr@{}}
    \toprule
    Method & RMSE$\downarrow$ & MAE$\downarrow$ & $r\uparrow$ & BIAS \\
    \midrule
    \multicolumn{5}{@{}l}{\textit{Forecast window: $(0,24]$ h ($N=2671$)}} \\
    \textbf{AeroJEPA} & \textbf{28.084} & \textbf{14.789} & \textbf{0.3340} & \textbf{+0.037} \\
    3D-Var + WRF  & 35.203 & 20.216 & 0.1248 & -7.989 \\
    Original WRF  & 36.809 & 21.946 & 0.1035 & -5.219 \\
    Persistence   & 37.570 & 22.591 & 0.0090 & -1.678 \\
    \midrule
    \multicolumn{5}{@{}l}{\textit{Forecast window: $(24,48]$ h ($N=2653$)}} \\
    \textbf{AeroJEPA} & \textbf{41.181} & 24.166 & \textbf{0.1160} & \textbf{+0.185} \\
    3D-Var + WRF  & 46.096 & \textbf{23.893} & -0.0541 & -11.422 \\
    Original WRF  & 46.309 & 24.309 & -0.0511 & -9.892 \\
    Persistence   & 47.221 & 25.903 & -0.1096 & -1.621 \\
    \midrule
    \multicolumn{5}{@{}l}{\textit{Forecast window: $(0,48]$ h ($N=5324$)}} \\
    \textbf{AeroJEPA} & \textbf{35.224} & \textbf{19.462} & \textbf{0.1937} & \textbf{+0.111} \\
    3D-Var + WRF  & 40.994 & 22.048 & 0.0251 & -9.700 \\
    Original WRF  & 41.814 & 23.123 & 0.0201 & -7.547 \\
    Persistence   & 42.653 & 24.242 & -0.0575 & -1.650 \\
    \bottomrule
  \end{tabular*}
\end{table}

\Cref{tab:pm25_multistep_results} summarizes the forecast-window and
full-period results. During the first 24 h, AeroJEPA reduces RMSE and
MAE by 20.22\% and 26.85\%, respectively, relative to 3D-Var + WRF.
Over the full 48 h, its RMSE and MAE are 35.224 and 19.462,
corresponding to reductions of 14.08\% and 11.73\%.
During the second 24 h, RMSE remains lower than that of 3D-Var + WRF,
whereas MAE is slightly higher.
In this case, AeroJEPA reduces full-period aggregate errors without
new observations, although its advantage weakens at longer lead times.

\begin{figure*}[!t]
  \centering
  \includegraphics[width=\textwidth]{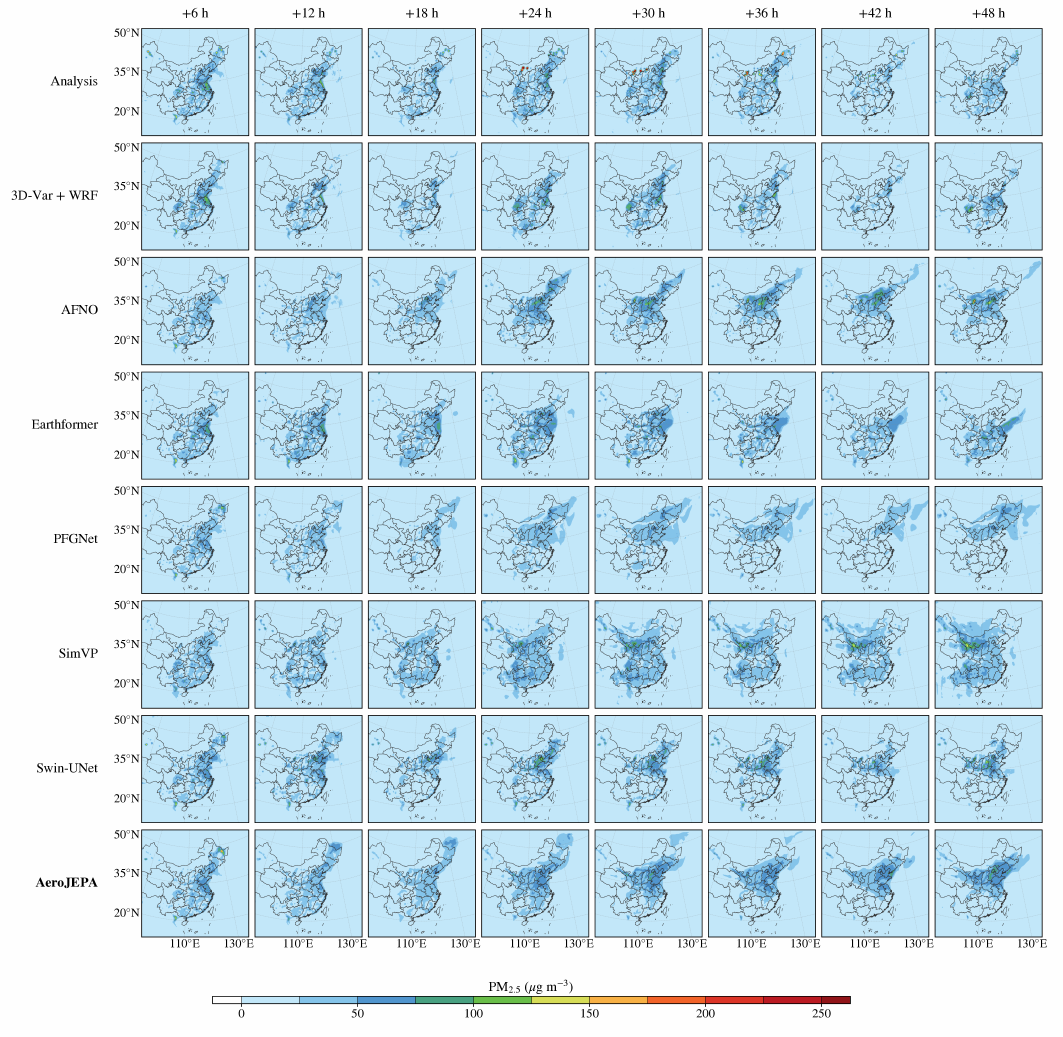}
  \caption{Spatial PM$_{2.5}$ distributions during the 48 h multi-step
  forecast. Units: $\mu\mathrm{g}\,\mathrm{m}^{-3}$.}
  \label{fig:supp_multistep_distribution}
\end{figure*}

The deep learning baselines in \cref{fig:supp_multistep_distribution}
use the same recursive protocol as AeroJEPA: predicted PM$_{2.5}$
and meteorological fields are fed back at each step, with observations
in newly added frames set to zero and their masks marked invalid.
No future observations or ground-truth meteorology are used.


MCVD is retained in the single-step benchmark. In this recursive
experiment, our task-adapted MCVD became unstable when its generated
PM$_{2.5}$ and meteorological fields were fed back as inputs; its saturated
concentration maps are therefore omitted from
\cref{fig:supp_multistep_distribution}.


\Cref{fig:supp_multistep_distribution} shows the 48 h recursive
forecasts. At longer lead times, 3D-Var + WRF retains large-scale
patterns similar to Analysis but misses some localized pollution
features and intensity changes. The deep learning models develop
different spatial displacements and morphological changes under
recursive feedback. Visually, AeroJEPA is closest to Analysis at +6 h,
but noticeable deviations emerge from +12 h onward, showing that
short-term accuracy does not ensure stable long-term predictions.


Despite these spatial deviations, AeroJEPA retains relatively
favorable independent-station metrics for some subsequent lead times.
Gridded evaluation measures spatial agreement with Analysis, whereas
station metrics assess concentration errors at monitoring locations.
Good station scores therefore need not imply spatial consistency
across the entire field; the two evaluations provide complementary
perspectives.


This case suggests that representations learned from historical
PM$_{2.5}$ and meteorology support short-term forecasting but may not
maintain realistic spatial evolution under prolonged recursive
extrapolation. Numerical models' physical constraints and ability to
preserve large-scale structures remain valuable. These results do not
support replacing those priors with learned representations;
incorporating numerical-model constraints may instead improve
long-term stability.


\subsection{Ablation Results}
\label{sec:supp_station_ablation}

\Cref{tab:supp_station_ablation} reports single-step ablation results
at the held-out stations, with variant names consistent with the
main paper.


\begin{table}[H]
  \centering
  \caption{Independent-station ablation results. RMSE, MAE, and BIAS
  are in $\mu\mathrm{g}\,\mathrm{m}^{-3}$; $r$ denotes Pearson correlation.
  Best results are bold; BIAS is best when closest to zero.}
  \label{tab:supp_station_ablation}
  \footnotesize
  \setlength{\tabcolsep}{1.5pt}
  \renewcommand{\arraystretch}{1.08}
  \begin{tabular*}{\linewidth}{@{\extracolsep{\fill}}lrrrr@{}}
    \toprule
    Model & RMSE$\downarrow$ & MAE$\downarrow$ & $r\uparrow$ & BIAS \\
    \midrule
    AeroJEPA & 15.60549 & 8.92855 & 0.71447 & \textbf{0.05211} \\
    w/o JEPA & 15.95015 & 8.90230 & 0.69911 & -0.32436 \\
    w/o CFF & \textbf{15.34923} & \textbf{8.86090} & \textbf{0.72567} & -0.20472 \\
    w/o CMG & 15.42208 & 8.96433 & 0.72259 & 0.36254 \\
    w/o MGP & 15.93945 & 9.00602 & 0.69962 & 0.34400 \\
    \bottomrule
  \end{tabular*}
\end{table}

The full model has the smallest absolute BIAS; w/o CFF leads in
RMSE, MAE, and correlation. Different rankings may reflect reference
and spatial sampling differences between 3D-Var analysis fields and
independent observations.


\setlength{\dbltextfloatsep}{8pt}
\vspace{-6pt}
\section{More Qualitative Results}
\label{sec:supp_qualitative}

\begin{figure*}[!t]
  \centering
  \includegraphics[width=\textwidth]{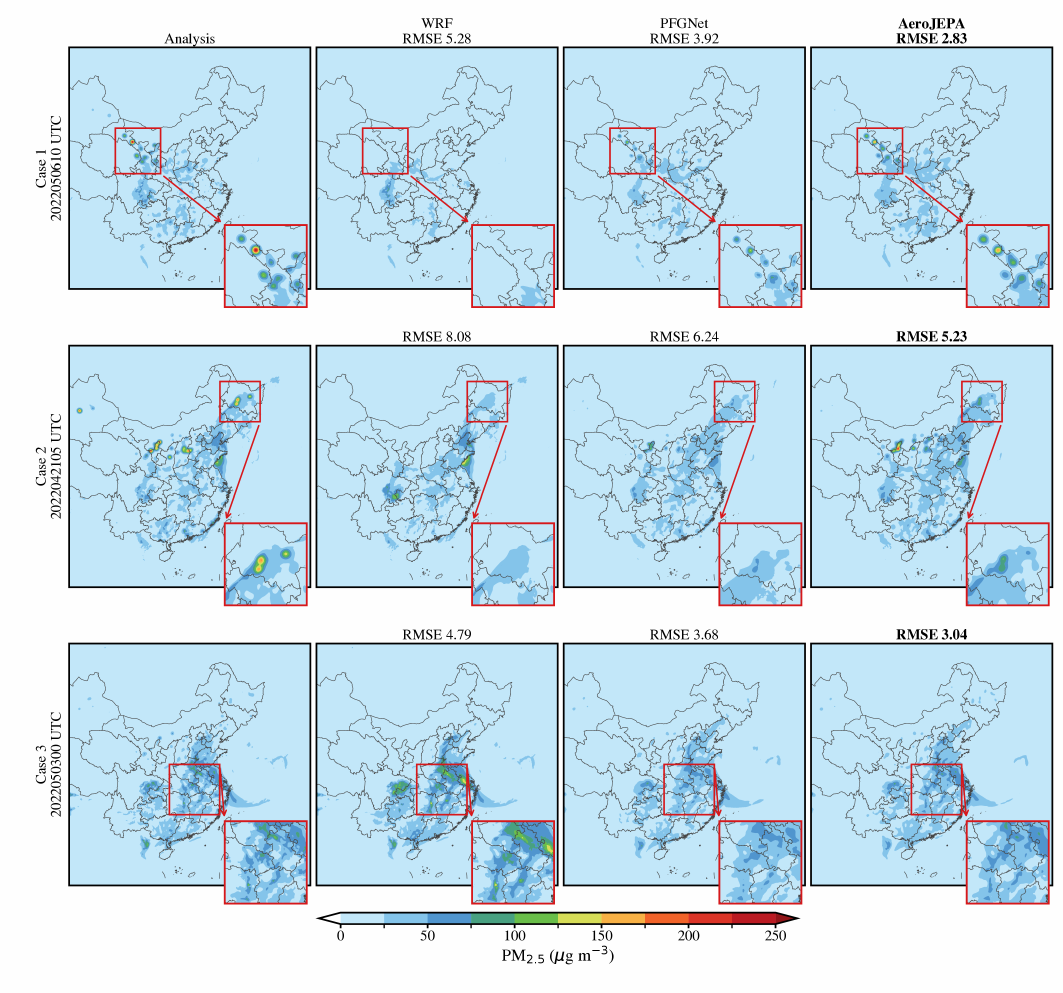}
  \caption{Additional +3 h PM$_{2.5}$ forecast comparisons with local
  enlargements. Concentrations and RMSE are in
  $\mu\mathrm{g}\,\mathrm{m}^{-3}$.}
  \label{fig:supp_qualitative_cases}
\end{figure*}
\Cref{fig:supp_qualitative_cases} shows three +3 h cases with local
enlargements. RMSE is computed over the full $400\times400$ grid
against Analysis.


\begin{samepage}
In Cases 1 and 2, WRF and PFGNet do not fully recover the localized
high-concentration features in northwestern and northeastern China,
respectively. AeroJEPA reproduces their spatial patterns more closely
than these baselines.
\end{samepage}


\begin{samepage}
In Case 3, WRF overestimates PM$_{2.5}$ over the middle and lower
Yangtze. PFGNet and AeroJEPA reduce this bias, with AeroJEPA closer
to Analysis. These cases illustrate the potential of historical
observations for correcting background biases and recovering local
concentration patterns.
\par
\end{samepage}


\end{document}